\documentclass[10pt,twocolumn,letterpaper]{article}

\usepackage{cvpr}
\usepackage{times}
\usepackage{epsfig}
\usepackage{graphicx}
\usepackage{amsmath}
\usepackage{amssymb}
\usepackage{multirow}

\usepackage[breaklinks=true,bookmarks=false]{hyperref}

\cvprfinalcopy 

\def\cvprPaperID{16} 

\ifcvprfinal\fi
\begin{document}

\title{Improved Soccer Action Spotting using both Audio and Video Streams}

\author{Bastien Vanderplaetse\\
ISIA Lab\\
University of Mons, Belgium\\
{\tt\small bastien.vanderplaetse@umons.ac.be}
\and
Stéphane Dupont\\
ISIA Lab\\
University of Mons, Belgium\\
{\tt\small stephane.dupont@umons.ac.be}
}

\maketitle

\begin{abstract}
In this paper, we propose a study on multi-modal (audio and video) action spotting and classification in soccer videos. Action spotting and classification are the tasks that consist in finding the temporal anchors of events in a video and determine which event they are. This is an important application of general activity understanding. Here, we propose an experimental study on combining audio and video information at different stages of deep neural network architectures. We used the SoccerNet benchmark dataset, which contains annotated events for 500 soccer game videos from the Big Five European leagues. Through this work, we evaluated several ways to integrate audio stream into video-only-based architectures. 
We observed an average absolute improvement of the mean Average Precision (mAP) metric of $7.43\%$ for the action classification task and of $4.19\%$ for the action spotting task.
\end{abstract}

\section{Introduction}\label{sec:introduction}
The annual revenue of the global sports market was estimated to $\$90.9$ billions in 2017~\cite{Statista.market.sports}. From this large amount, $\$28.7$ billion came from the European soccer market~\cite{Statista.market.sports.european}, more than half ($\$15.6$ billion) of which was generated by the Big Five European soccer leagues (EPL, Ligue 1, Bundesliga, Serie A and La Liga)~\cite{Statista.market.sports.european.bigfive,Statista.market.sports.european.top}. The main interest of sports broadcast is entertainment, but sports videos are also used by professionals for strategy analysis, player scouting or statistics generation. These statistics are traditionaly gathered by professional analysts watching a lot of videos and identifying the events occuring within a game. For football, this annotation task takes over 8 hours to provide up to 2000 annotations per game, according to Matteo Campodonico, \textsc{CEO} of Wyscout, a company specialized in soccer analytics~\cite{wyscout}. 

To assist sports annotators in this task, several automated computer vision methods can be devised to address many of the challenges in sports video understanding: field and lines localization~\cite{Farin2003RobustCC,Homayounfar2017SportsFL,Jiang2019OptimizingTL}, ball position~\cite{Kamble2017BallTI,Sarkar2019GenerationOB,Theagarajan2018SoccerWH} and camera motion~\cite{Lu2019PantiltzoomSF,Yao2016RobustMC} tracking, detection of players~\cite{Cioppa2019ARTHuSAR,Komorowski2019FootAndBallIP,Huda2018EstimatingTN}, their moves~\cite{Felsen2017WhatWH,Manafifard2017ASO,Thinh2019AVT}, and pose~\cite{Bridgeman2019MultiPerson3P,Zecha2019RefiningJL} and the team they are playing for~\cite{Istasse2019AssociativeEF}. Detecting key actions in soccer videos remains a difficult challenge since these events within the videos are sparse, making machine learning on massive datasets difficult to achieve. Some work has nevertheless achieved significant results in that direction~\cite{Cioppa2019ACL,Giancola_2018_CVPR_Workshops}.

In this paper, we focus on action spotting and classification in soccer videos. This task has been defined as finding the anchors of human-induced soccer events in a video~\cite{Giancola_2018_CVPR_Workshops} as well as naming the action categories. Several issues arise when dealing with this task. Important actions often have no clear start and end frames in the video, they are temporally discontinuous (i.e. adjacent frames may have different annotations), and they are rather rare. To improve action spotting performance, we propose to use both audio and video input streams while previous work did only use video. Different audio-visual neural network architectures are compared. Our intuition leads us to believe that some categories of actions trigger particular reactions on the part of the public present in the stadium. For example, when a goal is scored, fans shout out. Similarly, a red card can cause discontent. Audio signals should hence provide useful information in such key cases, for instance to distinguish real scored goals from goal attempts. This is what we will show in the paper.

\paragraph{Contributions.} \textbf{(i)} We carried out an initial analysis about the possibilities of adding audio as an input in a soccer action spotting and classification context. \textbf{(ii)} Our best approach improved the performance of action classification on SoccerNet~\cite{Giancola_2018_CVPR_Workshops} by $7.43\%$ absolute with the addition of audio, compared to the video-only baseline. \textbf{(iii)} We also increased the performance of the action spotting on the same dataset by $4.19\%$ absolute. 

\section{Related Work}
\paragraph{Sports Analytics and Related Applications.} Computer vision methods have been developed to help understand sport broadcasts, carry out analytics within a game~\cite{Corscadden2018DevelopingAT,DOrazio2010ARO,Thomas2017ComputerVF}, or even assist in broadcast production. Interesting use cases innclude the automatic summarization of games~\cite{Ekin2003AutomaticSV,10.1145/3347318.3355524,10.1145/3347318.3355526}, the identification of salient game actions~\cite{Feichtenhofer2016SpatiotemporalRN,Martnez2019ActionRW,Yaparla2019ANF} or the reporting of commentaries of live game video streams~\cite{Yu2018FineGrainedVC}.

Early work used camera shot segmentation and classification to summarize games~\cite{Ekin2003AutomaticSV} or focused on identifying video production patterns in order to detect salient actions of the game~\cite{Ren2005FootballVS}. Later, Bayesian networks have been used to detect goals, penalties, corner kicks and cards events~\cite{Huang2006SemanticAO} or to summarize games~\cite{Tavassolipour2014EventDA}.

More recently, deep learning approaches have been applied. Long Short-Term Memory (\textsc{LSTM}) networks~\cite{10.1162/neco.1997.9.8.1735} enabled to temporally traverse soccer videos to identify the salient actions by temporally aggregating particular features~\cite{Tsunoda2017FootballAR}. These features can be local descriptors, extracted by a Bag-of-Words (\textsc{BOW}) approach, or global descriptors, extracted by using Convolutional Neural Networks (\textsc{CNN}). Besides features, semantic information, such as player localization~\cite{Khan2018SoccerED}, as well as pixel information~\cite{Cioppa2018ABA}, are also used to train attention models to extract relevant frame features. Besides, a loss function for action spotting was proposed to tackle the issue of unclear action temporal location, by better handling the temporal context around the actions during training~\cite{Cioppa2019ACL}.

Some of the most recent works propose to identify kicks and goals in soccer games by using automatic multi-camera-based systems~\cite{Zhang2019AnAM}. Another work uses logical rules to define complex events in soccer videos in order to perform visual reasoning on these events~\cite{Khan2019VisualRO}. 
These complex events can be visualized as a succession of different visual observations during the game. For example, a \textit{``corner kick''} occurs when a player of the defending team hits the ball, which passes over the goal line. This complex event is the succession of simple visual observations: the ball is seen near a flag, a player comes near the position of the ball, this player kicks the ball, and the goal post becomes visible in the scene. The logical rules used to define these complex events are Event Calculus (\textsc{EC})~\cite{ec1}, i.e. a logic framework for representing and reasoning about events and their effects. These \textsc{EC} allow to describe a scene with atomic descriptions through First Order Logic (\textsc{FOL}).

\paragraph{Activity Recognition.} Activity recognition is the general problem of detecting and then classifying video segments according to a predefined set of activity or action classes in order to understand the videos. Most methods use temporal segments~\cite{Buch2017SSTST,Gao2017TURNTT,Yang2019ExploringFS} that need to be pruned and classified~\cite{Girdhar2017ActionVLADLS,Tang2019VideoSC}.

A common way to detect activities is to aggregate and pool these temporal segments, which allows to search for a consensus~\cite{Agethen2019DeepMC,Tran2019VideoCW}. Naive methods use average or maximum pooling, which require no learning. More complex ones aim to find a structure in a feature set by clustering and pooling these features while improving discrimination. These work use learnable pooling like \textsc{BOW}~\cite{Arandjelovic2013AllAV,Jgou2010AggregatingLD}, Fisher Vector~\cite{Darczy2013FisherKF,Nagel2015MetaFV,Pan2019ForegroundFV} or \textsc{VLAD}~\cite{Arandjelovic2013AllAV}. Some works improve these techniques by respectively extending them by the incorporation of the following Deep Neural Network (\textsc{DNN}) architectures: Net\textsc{FV}~\cite{Tang2019DeepFF}, Soft\textsc{DBOW}~\cite{Philbin2008LostIQ} or Net\textsc{VLAD}~\cite{Girdhar2017ActionVLADLS}.

Instead of pooling features, some works try to identify which features might be the more useful given the video context. Some of these approaches represent and harness information in both temporal and/or spatial neighborhoods~\cite{Dai2017TemporalCN,Liu2019MultiScaleBC}, while other ones focus on attention models~\cite{Nguyen2015STAPSA,Wang2018FastAA} as a way to better leverage the surrounding information by learning adaptive confidence scores. For instance, the evidence of objects and scenes within a video can be exploited by a semantic encoder for improving activity detection~\cite{Heilbron2017SCCSC}. Moreover, coupling recognition with context-gating allows the learnable pooling methods to produce state-of-the-art recognition performance on very large benchmarks~\cite{DBLP:journals/corr/MiechLS17}.

Advanced methods for temporal integration use particular neural network architectures, such as Convolution Neural Network (\textsc{CNN})~\cite{Shou2016TemporalAL} or Recurrent Neural Network (\textsc{RNN})~\cite{Pei2016TemporalAM}. More particularly, \textsc{LSTM} architectures are often chosen for motion-aware sequence learning tasks, which is beneficial for activity recognition~\cite{Agethen2019DeepMC,Baccouche2010ActionCI}. Attention models are also harnessed to better integrate spatio-temporal information. Within this category of approaches, recent work uses a 2-models-based attention mechanism~\cite{Peng2019TwoStreamCL}. The first one consists of a spatial-level attention model, which determines the important regions in a frame, and the second one concerns the temporal-level attention, which is used to harness the discriminative frames in a video. Another work proposes a convolutional \textsc{LSTM} network supporting multiple convolutional kernels and layers coupled with an attention-based mechanism~\cite{Agethen2019DeepMC}.

\paragraph{Multimodal approaches.} Using several different and complementary input modalities can improve model performance in both action classification and action spotting tasks, since this leverages more information about the video. Earlier work uses textual sources~\cite{Oskouie2012MultimodalFE}, such as the game logs manually encoded by operators.

Recently, research in multimodal models use, in addition to the RGB video streams, information about the motion within the video sequences: the optical flow can be used~\cite{Ye2019TwoStreamCN,Yudistira2020CorrelationNS} or even player pose sequences~\cite{Cai2018TemporalHA,Vats2019TwoStreamAR}. For golf and tennis tournaments, a multimodal architecture using the reactions (such as high fives or fist pumps) and expressions of the players (aggressive, smiling, etc.), spectators (crowd cheering) and commentator (tone and word analysis), and even game analytics, was proposed~\cite{Merler2019AutomaticCO}.

Some work use the audio stream of the video but in a different manner than ours. The audio stream was used to make audio-visual classification of sport types~\cite{Gade2015AudioVisualCO}. Also, acoustic information was used to detect tennis events and track time boundaries of each tennis point in a match~\cite{Baughman2019DetectionOT}.

\section{Methodology}
Our main objective is to set-up multimodal architectures and analyze the benefit of the audio stream on the performance of a model within the soccer action spotting and classification tasks. Action spotting is the task that consists in finding the right temporal anchors of events in a video. The more a candidate spot is close to the target event, the more the spotting is considered as good. Reaching perfect spotting is hence particularly complex. Tolerance intervals are hence typically used.Regarding classification, we will use a typology of different soccer actions classes and evaluate how well our systems distinguish those classes.

We use the SoccerNet dataset~\cite{Giancola_2018_CVPR_Workshops}. It uses a typology of 3 soccer event categories: \textit{goals}, \textit{substitutions} and \textit{cards} (both yellow and red cards).

This section starts by explaining how the video and the audio streams are represented with feature vectors to be used as input of the different models. Next, it presents the baseline approach proposed in \cite{Giancola_2018_CVPR_Workshops}. This approach consists of training models for soccer action classification but to include a background class too, so that both classification and action spotting tasks can be addressed.Since the baseline \cite{Giancola_2018_CVPR_Workshops} uses only video stream, we finish this section by exposing how we use the audio stream too, with different variants for multimodal fusion.

\subsection{Video and Audio Representations}\label{subsec:front-end}
We want to work with both video and audio streams. The volume of data available for training may however be insufficient for training fully end-to-end machine learning architectures. Hence, we will here reuse existing visual and auditory feature extraction architectures, pre-trained on relevant visual and audio reference datasets. As explained in more details later, we used a ResNet~\cite{DBLP:journals/corr/HeZRS15} trained on the ImageNet~\cite{imagenet} data for the visual stream and, for the audio stream, a \textsc{VGG}~\cite{vgg} trained on spectrogram representations of the AudioSet~\cite{audioset} data. Fine-tuning of these models might be considered in the future, but at this stage, we keep their parameters fixed during our training process. In practice, we hence extracted visual and auditory features before running our experiments.

\paragraph{Video streams.} For the video streams, we used the features extracted by~\cite{Giancola_2018_CVPR_Workshops} using ResNet-152~\cite{DBLP:journals/corr/HeZRS15}, a deep convolutional neural network, pretrained on 1000 categories ImageNet~\cite{imagenet} dataset. Particularly, they used the output of the \textit{fc1000} layer, which is a 1,000-way fully-connected layer with a softmax function in the end. This layer outputs a 2,048-dimensional feature vector representation for each frame of the video. To extract these features, each video was unified at $25$ frames per second (fps) and trimmed at the start of the game, since the reference time for the event annotations is the game start. Each frame was resized and cropped to a $224 \times 224$ resolution. A TensorFlow~\cite{DBLP:journals/corr/AbadiABBCCCDDDG16} implementation was used to extract the features of the videos every 0.5 second, hence a 2 frames per second sampling rate. Then, Principal Component Analysis (\textsc{PCA}) was applied on the extracted features to reduce their dimension to 512. This still retain $93.9\%$ of their variance\footnote{Although the reference publication does not mention it, we assume the PCA transformation matrix is estimated on the SoccerNet training data.}

\paragraph{Audio streams.} For the audio streams, we used \textsc{VGG}~\cite{vgg}, a deep convolutional network architecture. Particularly, we used \textsc{VGG}ish, a \textsc{VGG} architecture pretrained on AudioSet~\cite{audioset}. AudioSet is a benchmark dataset containing 632 audio event categories and 1,789,621 labeled 10-seconds audio segments from YouTube videos. To extract audio features, we used a TensorFlow implementation of a pretrained slim version of \textsc{VGG}ish\footnote{\href{https://github.com/DTaoo/VGGish}{https://github.com/DTaoo/VGGish}}. We extracted the output of the last convolutional layer (\textit{conv4/conv4\_2}) to which we applied a global average pooling to get 512-dimensional feature vector. Since this model uses a chunk of the audio stream spectrogram as input and we want to have the same frame rate as the video features, we trimmed the audio streams at the game start and divided them into chunks of 0.5 second.

\subsection{SoccerNet baseline approach}\label{subsec:baseline}
The baseline approach proposed in~\cite{Giancola_2018_CVPR_Workshops} is divided into two parts: \textbf{(i)} video chunk classification and \textbf{(ii)} action spotting. In order to compare the performance with and without the audio stream, we followed the same approach and used the best performing models as baselines for our approach.

\paragraph{Video chunk classification.} For the classification task, shallow pooling neural networks are used. Each video is chunked into windows of duration $T$ seconds. Since the features are sampled at 2 frames per second, the input matrix to our systems is, for each chunk to be classified, an aggregation of $W=2T$ feature vectors. Therefore, the dimension of the input is $W \times 512$. Although quite rare, some chunks may have multiple labels when several actions are temporally close-by. Our deep learning architectures will hence use a sigmoid activation function at their last layer. For all the classes, a multi binary cross-entropy loss is minimized. The Adam optimizer is used. The learning rate follows a step decay and early stopping is applied, based on the validation set performance. The final evaluation metric used is the mAP accross the classes defined on SoccerNet~\cite{Giancola_2018_CVPR_Workshops}.

One of the main challenges in designing the neural network architectures for this task was related to the temporal pooling method to be used. Indeed, the selected feature extraction approaches use fixed image based models, while we want to use chunks consisting of several video frames as input to provide the system with longer term information that should be beneficial (or even necessary) to achieve useful performance. 

Seven pooling techniques have been tested by~\cite{Giancola_2018_CVPR_Workshops}: \textbf{(i)} mean pooling and \textbf{(ii)} max pooling along the aggregation axis outputting 512-long features, \textbf{(iii)} a custom \textsc{CNN} with kernel dimension of $512 \times 20$ traversing the temporal dimension. At last, the approaches and implementations proposed by~\cite{DBLP:journals/corr/MiechLS17} such as \textbf{(iv)} Soft\textsc{DBOW}, \textbf{(v)} Net\textsc{FV}, \textbf{(vi)} Net\textsc{VLAD} and \textbf{(vii)} Net\textsc{RVLAD} have also been compared. These last pooling methods use clustering-based aggregation techniques to harness context-gating, which is a learnable non-linear unit aiming to model the interdependencies among the network activations. To predict the labels for the input window, a fully connected layer was then stacked after the pooling layer of each model. Dropout is also used during training in order to improve generalization. We use a keep probability of $60\%$.

\paragraph{Action spotting.} For the spotting task, \cite{Giancola_2018_CVPR_Workshops} reused their best performing model from the classification task. This model is applied on each testing video. In this case, instead of splitting video into consecutive chunks, a sliding window of size $W$ is used through the videos, with a stride of 1 second.
Therefore, for each position of the sliding window, a $W \times 512$ matrix can be obtained by the content currently covered by this window. This matrix is used as input for the model, which computes a probability vector for classifying the video chunk. Then, we get for each game a series of predictions consisting of probabilities to belong to each class (including the background no-action class).

To obtain the final spotting candidates from the predictions series, three methods were used in \cite{Giancola_2018_CVPR_Workshops}: \textbf{(i)} a watershed method using the center time within computed segment proposals; \textbf{(ii)} the time index of the maximum value of the watershed segment as the candidate; and \textbf{(iii)} the local maxima along the video and applying non-maximum-suppression (\textsc{NMS}) within the window. Here, a tolerance $\delta$ is added to the mAP as evaluation metric. Therefore, a candidate spot is defined as positive if it lands within a tolerance of $\delta$ seconds around the true temporal anchor of an event. Another metric is the Average-mAP, which is the area under the mAP curve with $\delta$ ranging from 5 to 60 seconds.

In this paper, we use Net\textsc{VLAD} and Net\textsc{RVLAD} as pooling layers, since they are the best approaches in the chosen baseline.

\subsection{Audio Input and Multimodal Fusion}
We need to define architectures of audio-visual models in order to study the influence of the audio stream on performance. We decided to use the baseline architecture described in Section \ref{subsec:baseline} for both visual and audio streams, the only difference being the feature extraction front-end (cfr. \ref{subsec:front-end}).

\begin{figure*}
\begin{center}
   \includegraphics[width=1.0\linewidth]{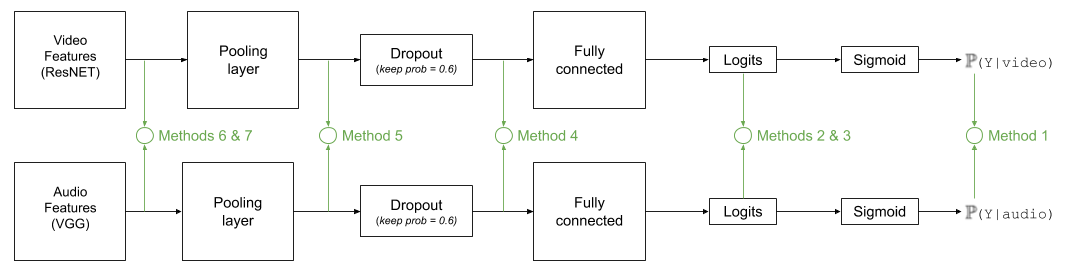}
\end{center}
   \caption{Example of a short caption, which should be centered.}
\label{fig:pipeline}
\end{figure*}

Next, we need to determine where the two models need to merge their pipeline in order to get better results. Both visual and audio processing pipelines can be applied until their last layers, with their outputs then being combined with a late fusion mechanism. Earlier fusion points will also be investigated. The general appearance of our multi-modal pipeline is illustrated in Figure \ref{fig:pipeline}. We hence train our models with different \textit{merge points}, illustrated by green circles and arrows on the figure. At the merge points, the audio stream and video stream representation vectors are concatenated, and the concatenated vectors are then processed by a single pipeline consisting of the remaining part of the baseline processing pipeline downstream the fusion point. We distinguish 5 merge points and 7 methods.

The first two methods are the only ones to be applied after having trained both models. The first method multiplies the probabilities estimated for each class, while the second one averages the logits, followed by the sigmoid activation function, applied on the resulting logits vectors. The third method applies the same process then method one, but the two parallel models are trained with a loss computed from the output of the common sigmoid function, instead of using pre-trained models. Methods 4 and 5 have their merge point respectively before the fully-connected layer, and before the dropout layer of the model. Merging before and after the dropout can lead to different results. Indeed, if we merge after the dropout, we will keep for each training sample the same proportion of activations from both flows, while merging before the dropout does not ensure that the information from both streams is kept fairly. Methods 6 and 7 both merge the representation vectors before the pooling layer. The difference between these methods lies in the size of the pooling layer output (1,024 for method 5 and 512 for method 6).

Every method, except the two first ones, requires a training since there are learnable parameters in both the pooling process and in the fully-connected layers.

\section{Experiments}
\subsection{SoccerNet dataset}
For our experiments, we use the same dataset as our reference baseline: SoccerNet~\cite{Giancola_2018_CVPR_Workshops}. This dataset contains videos for 500 soccer games from the Big Five European leagues (\textsc{EPL}, La Liga, Ligue 1, Bundesliga and Serie A): 300 as training set, 100 as validation set and 100 as testing set. There are 6,637 events referenced for these games, split into 3 classes: \textit{``goals''}, the instant the ball crosses the goal line to enter the net; \textit{``cards''}, the instant a yellow or a red card is shown by the referee; and \textit{``substitutions''}, the instant a new player enters in the field to replace another one. Each one of these events is annotated by the exact second it occurs in the game. For the classification task, a fourth class was added: \textit{``background''}, which corresponds to the absence of the three events.

\subsection{Video chunk classification}\label{subsec:expclf}
We train models with our 7 fusion methods. In the baseline, the best model uses Net\textsc{VLAD} as pooling layer, with a number of clusters of $k=512$. However, such a large number of clusters incurs a larger computational load, which increases linearly with the value of $k$. Therefore, we first compare our merging methods with a smaller number of clusters: $k=64$. According to~\cite{Giancola_2018_CVPR_Workshops}, the best pooling method with a number of cluster $k = 64$ is Net\textsc{RVLAD}. Therefore, we try our merging methods on models having a $64$-clusters Net\textsc{RVLAD} as pooling layer. Our results are presented on Table~\ref{tab:rvladclassification}. The video baseline result is obtained by executing the code provided by Giancola et a., and the audio baseline uses the same code, but using the audio stream as input. We also compared the performance for chunks of size 60 seconds or else 30 seconds.

We observe that a chunk of 60 seconds provides better results. This can be explained by the fact that, with 30-seconds video chunks, the \textit{``background''} class represent $93\%$ of the training data samples, whereas for a 60-seconds window, it represents $87\%$. Since there are more samples in the \textit{``background''} class, the 30-seconds models tend to classify more samples with this label, which reduces performance on other classes.

Regarding multimodal fusion, we can see that using only the audio stream provides inferior results than the video-only model. On the other hand, all our methods to combine video and audio streams improve over the performance of mono-modal systems. The best performance is obtained by the fourth merging method, which correspond to the merge point localized before the last fully connected classification layer.

In Table~\ref{tab:scorelabel}, we compare the mAP for each class for the video baseline, the audio baseline and our best fusion method. We can observe that including audio improves the performance on each category, especially for the \textit{``goals''} event, where the relative reduction of the error rate exceeds 50\%. Moreover, if the audio baseline generally performs worse than the video baseline, this is not the case for the \textit{``goals''} class, where audio alone yields better results than video alone. This corroborates our intuitions exposed in Section~\ref{sec:introduction}. Indeed, a scored goal, which clearly leads to a strong emotional reaction from the public as well as the commentators, is easier to detect through the audio stream than the video stream, where it could for instance be confused with shots on target. However, the audio stream does not seem to provide sufficient information to efficiently detect cards, leading to a poor result for this category. Finally, audio carries information about the substitutions. This can likely be explained by the fact that the public can applaud or boo the player that comes in or comes out of the field, depending on his status or the quality of his play during the game.

\begin{table}
\caption{Classification metric (mAP) for different merging methods and different video chunk sizes using Net\textsc{RVLAD}, with $k=64$ clusters, as pooling layer.}
\label{tab:rvladclassification}
\begin{center}
    \begin{tabular}{c|c|c}
\textbf{Models}                                                                           & \textbf{$T = 60$ sec.} & \textbf{$T = 30$ sec.} \\ \hline
\textbf{Video baseline \cite{Giancola_2018_CVPR_Workshops}} & 66.0                      & 58.7                      \\
\textbf{Audio baseline}                                                                   & 50.6                      & 43.7                      \\
\textbf{Merging method 1}                                                                   & 68.4                      & 63.7                     \\
\textbf{Merging method 2}                                                                   & 72.6                      & 67.3                      \\
\textbf{Merging method 3}                                                                   & 73.4                      & 69.3                      \\
\textbf{Merging method 4}                                                                   & \underline{73.7}                      & 68.8                      \\
\textbf{Merging method 5}                                                                   & 72.8                      & 68.7                      \\
\textbf{Merging method 6}                                                                   & 64.1                      & 59.6                      \\
\textbf{Merging method 7}                                                                   & 64.2                      & 58.1                      
\end{tabular}
\end{center}
\end{table}

\begin{table}
\caption{Comparison of the classification metric (mAP) on each label.}
\label{tab:scorelabel}
\begin{center}
    \begin{tabular}{c|c|c|c|}
\textbf{\begin{tabular}[c]{@{}c@{}}Labels \\ \end{tabular}} & \textbf{\begin{tabular}[c]{@{}c@{}}Video \\ baseline \cite{Giancola_2018_CVPR_Workshops}\end{tabular}} & \textbf{\begin{tabular}[c]{@{}c@{}}Audio \\ baseline\end{tabular}} & \textbf{\begin{tabular}[c]{@{}c@{}}Merge \\ method 4\end{tabular}}\\ \hline
\textit{``background''} & 97.6 & 96.7 & 98.0 \\
 \textit{``cards''} & 60.5 & 19.2 & 63.9 \\
 \textit{``substitutions''} & 69.8 & 55.1 & 72.6 \\
 \textit{``goals''} & 67.7 & 77.3 & 84.5 
\end{tabular}
\end{center}
\end{table}

Another interesting observation concerns the difference between the confusion matrices of 60-seconds and 30-seconds models. Table~\ref{tab:confusion} presents these confusion matrices for the model trained with the merge point before the fully connected layer (fourth merging method). If we focus only on the samples classified in one of the three events of interest (\textit{``cards''}, \textit{``substitutions''} and \textit{``goals''}), we can see that the proportion of errors is lower in the 30-seconds version ($2.68\%$ instead of $4.83\%$). This observation can be explained by the fact that a smaller video chunk size reduces the probability to have multiple different events in the same window. Therefore, it becomes easier to determine the differences between the three classes of interest. However, as explained earlier, the overall mAP score is worse due to the higher proportion of \textit{``background''} samples.

\begin{table*}
\caption{Confusion matrix for the model using the fourth merging method, with 60-seconds video chunks and 30-seconds video chunks.}
\label{tab:confusion}
\begin{center}
\begin{tabular}{clcccc}
\multicolumn{6}{l}{\textbf{60-seconds video chunks}}                                                                                                                                                                                                                                                                 \\ \cline{3-6} 
\multicolumn{1}{l}{}                                                                                         & \multicolumn{1}{l|}{}                    & \multicolumn{4}{c|}{\textbf{Predicted labels}}                                                                                                            \\ \cline{3-6} 
\multicolumn{1}{l}{}                                                                                         & \multicolumn{1}{l|}{}                    & \multicolumn{1}{l|}{\textit{background}} & \multicolumn{1}{l|}{\textit{cards}} & \multicolumn{1}{l|}{\textit{subs}} & \multicolumn{1}{l|}{\textit{goals}} \\ \hline
\multicolumn{1}{|c|}{\multirow{4}{*}{\textbf{\begin{tabular}[c]{@{}c@{}}Groundtruth\\ labels\end{tabular}}}} & \multicolumn{1}{l|}{\textit{background}} & \multicolumn{1}{c|}{7673}                & \multicolumn{1}{c|}{95}            & \multicolumn{1}{c|}{80}           & \multicolumn{1}{c|}{42}             \\ \cline{2-6} 
\multicolumn{1}{|c|}{}                                                                                       & \multicolumn{1}{l|}{\textit{cards}}      & \multicolumn{1}{c|}{178}                  & \multicolumn{1}{c|}{243}            & \multicolumn{1}{c|}{13}             & \multicolumn{1}{c|}{3}              \\ \cline{2-6} 
\multicolumn{1}{|c|}{}                                                                                       & \multicolumn{1}{l|}{\textit{subs}}       & \multicolumn{1}{c|}{175}                  & \multicolumn{1}{c|}{9}             & \multicolumn{1}{c|}{310}           & \multicolumn{1}{c|}{11}              \\ \cline{2-6} 
\multicolumn{1}{|c|}{}                                                                                       & \multicolumn{1}{l|}{\textit{goals}}      & \multicolumn{1}{c|}{91}                  & \multicolumn{1}{c|}{1}              & \multicolumn{1}{c|}{2}            & \multicolumn{1}{c|}{215}            \\ \hline
\multicolumn{1}{l}{}                                                                                         &                                          & \multicolumn{1}{l}{}                     & \multicolumn{1}{l}{}                & \multicolumn{1}{l}{}               & \multicolumn{1}{l}{}                \\
\multicolumn{6}{l}{\textbf{30-seconds video chunks}}                                                                                                                                                                                                                                                                 \\ \cline{3-6} 
\multicolumn{1}{l}{}                                                                                         & \multicolumn{1}{l|}{}                    & \multicolumn{4}{c|}{\textbf{Predicted labels}}                                                                                                            \\ \cline{3-6} 
\multicolumn{1}{l}{}                                                                                         & \multicolumn{1}{l|}{}                    & \multicolumn{1}{l|}{\textit{background}} & \multicolumn{1}{l|}{\textit{cards}} & \multicolumn{1}{l|}{\textit{subs}} & \multicolumn{1}{l|}{\textit{goals}} \\ \hline
\multicolumn{1}{|c|}{\multirow{4}{*}{\textbf{\begin{tabular}[c]{@{}c@{}}Groundtruth\\ labels\end{tabular}}}} & \multicolumn{1}{l|}{\textit{background}} & \multicolumn{1}{c|}{16768}               & \multicolumn{1}{c|}{97}            & \multicolumn{1}{c|}{116}           & \multicolumn{1}{c|}{43}            \\ \cline{2-6} 
\multicolumn{1}{|c|}{}                                                                                       & \multicolumn{1}{l|}{\textit{cards}}      & \multicolumn{1}{c|}{224}                  & \multicolumn{1}{c|}{212}            & \multicolumn{1}{c|}{5}            & \multicolumn{1}{c|}{0}              \\ \cline{2-6} 
\multicolumn{1}{|c|}{}                                                                                       & \multicolumn{1}{l|}{\textit{subs}}       & \multicolumn{1}{c|}{217}                 & \multicolumn{1}{c|}{11}              & \multicolumn{1}{c|}{305}           & \multicolumn{1}{c|}{4}              \\ \cline{2-6} 
\multicolumn{1}{|c|}{}                                                                                       & \multicolumn{1}{l|}{\textit{goals}}      & \multicolumn{1}{c|}{114}                  & \multicolumn{1}{c|}{0}              & \multicolumn{1}{c|}{0}             & \multicolumn{1}{c|}{209}            \\ \hline
\end{tabular}
\end{center}
\end{table*}

After finding the best merging method, we used the best baseline model from~\cite{Giancola_2018_CVPR_Workshops}, i.e. with a $512$-clusters Net\textsc{VLAD} as pooling layer, and we trained it three times with different input configurations: \textbf{(i)} only video stream, \textbf{(ii)} only audio stream, and \textbf{(iii)} both video and audio stream, by using our best merging method. For each one of these configurations, we used 60-seconds and 20-seconds video chunks.

Table~\ref{tab:vladclassification} presents the mAP score for each of these models. As previously, we observe that using only audio stream performs worse than video stream alone but the combination of the two streams provides significantly improved performance. Moreover, the use of 60-seconds chunks for training performs way better than the use of 20-seconds chunks, except for the combination of audio and video, where the difference is non-significant.

\begin{table}
\caption{Classification metric (mAP) for models using Net\textsc{VLAD}, with $k=512$ clusters, as pooling layer.}
\label{tab:vladclassification}
\begin{center}
\begin{tabular}{c|c|c}
\textbf{Models}                                                                 & \textbf{$T = 60$ sec.} & \textbf{$T = 20$ sec.} \\ \hline
\textbf{\begin{tabular}[c]{@{}c@{}}Video-based\\ NetVLAD baseline\end{tabular}} & 67.5                   & 56.6                   \\ \hline
\textbf{\begin{tabular}[c]{@{}c@{}}Audio-based\\ NetVLAD baseline\end{tabular}} & 46.8                   & 35.9                   \\ \hline
\textbf{\begin{tabular}[c]{@{}c@{}}Audio + Video\\ NetVLAD\end{tabular}}        & 75.2                   & 75.0                  
\end{tabular}
\end{center}
\end{table}

The model using Net\textsc{VLAD} as pooling layer and both video and audio streams is the one registering the best results for the classification task with a mAP score of $75.2\%$. In average, adding the audio stream as input to the models increases the mAP by $7.43\%$ in absolute terms compared to the video-only-based models.

\subsection{Action spotting}
Following the methodology proposed by \cite{Giancola_2018_CVPR_Workshops}, the action spotting task, as described in Section~\ref{subsec:baseline}, uses the best trained models from the classification task and the spotting results are obtained using three method variants: segment center, segment maximum and \textsc{NMS}. For each method, we compute the Average-mAP, which is the area under the mAP curve as a function of a tolerance $\delta$ in the precise time instant of the detected event, ranging from 5 to 60 seconds. In order to make comparisons, we applied this spotting process to the 6 trained models using Net\textsc{VLAD} as pooling layer, and to 3 of the models using Net\textsc{RVLAD}: \textbf{(i)} video-only, \textbf{(ii)} audio-only, and \textbf{(iii)} both video and audio with the merge point before the fully connected layer. Table~\ref{tab:averagemap} presents the Average-mAP for each.

\begin{figure}[t]
\begin{center}
  \includegraphics[width=1.0\linewidth]{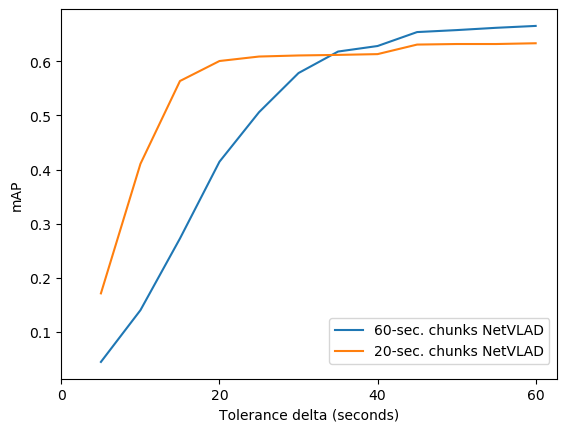}
\end{center}
  \caption{mAP score as a function of tolerance $\delta$ for Net\textsc{VLAD}-based models, trained with 20-seconds and 60-seconds video chunks, by using both audio and video streams.} 
\label{fig:tolerance}
\end{figure}

 Similarly to the classification task, using only audio is not as good as using video alone, but the combination improves the performance. What differs from classification is that smaller video chunks leads to better results, regardless of the method used. Our intuition is that shorter chunks enable to distinguish and detect actions that are temporally closer to each other. However, the high difference in the Average-mAP scores between models trained on 20-seconds windows and the ones trained on 60-seconds windows is particularly important. This can be explained by the fact that models trained with 60-seconds video chunks will have decreasing performances when tolerance $\delta$ becomes lower than $60$ seconds since the models was not trained for this. However, if models trained on 20-seconds video chunks performs well with $\delta=20$ seconds, it will still be efficient for higher values of $\delta$. Figure~\ref{fig:tolerance} illustrates this suggestions by showing the mAP as a function of tolerance $\delta$ for both Net\textsc{VLAD}-based models using audio and video streams. We can observe that both models tend to their best performance when $\delta$ is higher or equals to the corresponding window size.
 
The best model is the one trained on 20-seconds video chunks and using Net\textsc{VLAD} as pooling layer and both video and audio streams, yielding an Average-mAP of $56\%$. In average, adding the audio stream as input to the models increases the Average-mAP of $4.19\%$ absolute compared to the video-only models.

\setlength{\tabcolsep}{4pt}
\begin{table*}
\caption{Average-mAP for action spotting.}
\label{tab:averagemap}
\begin{center}
\begin{tabular}{c|c|c|c|c|c|c|c|c|c}
& \multicolumn{3}{c|}{\textbf{Video-only}} & \multicolumn{3}{c|}{\textbf{Audio-only}} & \multicolumn{3}{c}{\textbf{Audio + Video}} \\ \hline
\textbf{Models} & \textbf{Seg. max} & \textbf{Seg. center} & \textbf{NMS} & \textbf{Seg. max} & \textbf{Seg. center} & \textbf{NMS} & \multicolumn{1}{c|}{\textbf{Seg. Max}} & \multicolumn{1}{c|}{\textbf{Seg. center}} & \textbf{NMS} \\ \hline
\textbf{NetRVLAD}   
    & 30.8\%    & 41.9\%    & 30.2\%    & 21.8\%    & 30.3\%    & 22.1\%    & 34.0\%    & 47.6\%    & 33.4\%    \\ \hline
\textbf{\begin{tabular}[c]{@{}c@{}}60-sec. chunks\\ NetVLAD\end{tabular}} 
    & 29.6\%    & 43.4\%    & 29.0\%    & 19.9\%    & 27.1\%    & 19.5\%    & 32.3\%    & 48.7\%    & 31.8\%    \\ \hline
\textbf{\begin{tabular}[c]{@{}c@{}}20-sec. chunks\\ NetVLAD\end{tabular}}
    & 49.2\%    & \underline{50.2\%}  & 49.4\%    & 30.0\%    & \underline{31.0\%}  & 30.0\%    & 54.0\%    & \underline{56.0\%}  & 53.6\%    
\end{tabular}
\end{center}
\end{table*}
\setlength{\tabcolsep}{6pt}

\subsection{Additional observations}

\begin{figure}[t]
\begin{center}
  \includegraphics[width=1.0\linewidth]{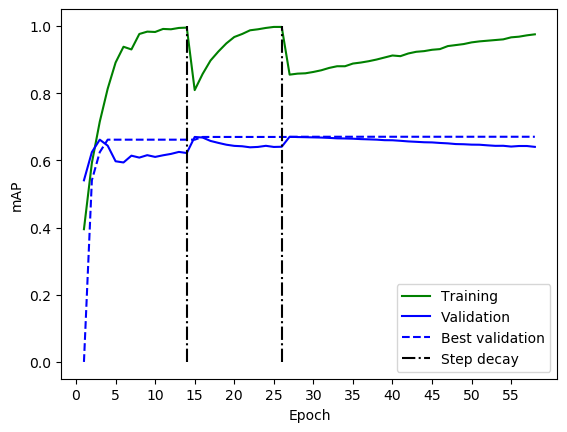}
\end{center}
  \caption{Evolution of the learning curves, i.e. training and validation mAP score curves, through the epochs for our best model.}
\label{fig:learningcurves}
\end{figure}

Figure \ref{fig:learningcurves} presents the evolution of performance on both the training set and the validation set during the training process, for our best model. The green line represents the evolution of the mAP classification score on the training set and the blue line is the evolution of the mAP score on the validation set. The horizontal dotted blue line represent the best mAP score reached on the validation set. The vertical black lines indicate the epochs at which a step decay was applied on the learning rate. 

We can observe that the mAP score on the training set quickly reaches a very high value, while performance on the validation set always remains much lower. A generalization gap of about  $35\%$ is visible, between the performance on the training set and on the validation set. Even our best performing model significantly overfits, possibly as a consequence of the still too small size of the training set. Indeed, despite being one of the best benchmark for the soccer action spotting and classification challenges, SoccerNet contains annotations for 500 games, which represents only 3,965 annotated events available for training.

This represents one of the current limitations of our study, and strategies to either increase the training set size, reduce over-fitting, or increase the generalization capabilities of our models should represent an important avenue for future research.

\section{Future Work}
In future works, we suggest to pursue the exploration of additional types of input streams, like optical flow, or even language streams such as transcriptions of commentators speech.

Furthermore, exploring more elaborate fusion mechanisms could be interesting. In order to improve our current models, one could also harness others feature extraction models than ResNet and VGG.

Another aspect that can be analyzed more in depth is trying to get a better understanding of the information carried by the audio stream. The audio in SoccerNet videos contain a mix between the commentators' voice and the sound coming from the stadium , including the field and the public. Therefore, we still do not know which of those different information source have the most impact on performance.

To address the issue related to the size of SoccerNet dataset, increasing the number of training samples is a possible solution. This could rely in part on data augmentation strategies, as well as annotating additional soccer game videos, or making use of unsupervised learning techniques.

\section{Conclusion}
In this paper, we studied the influence of the audio stream on soccer action classification and action spotting tasks, with performance evaluations on the SoccerNet baseline. For both tasks, using only the audio stream provides worse results than using only the video stream, except on the \textit{``goals''} class, where audio significantly exceeds video performance. Furthermore, combining both streams yields to better results on every category of actions. Combining audio and video streams improves, in average, the performance of action classification on SoccerNet by $7.43\%$ absolute, and the performance of action spotting by $4.19\%$. We also showed that using smaller video chunk sizes performs worse on classification, but improves the results for the action spotting task. 

A more in-depth study of the audio stream could lead to a better understanding of what actually provides information that the visual processing model fails to identify. In particular, separating the voices of commentators from the sound ambiance coming from the stadium could definitely help in this study.

{\small
\bibliographystyle{ieee_fullname}
\bibliography{egbib}

\begin{thebibliography}{10}\itemsep=-1pt

\bibitem{DBLP:journals/corr/AbadiABBCCCDDDG16}
Mart{\'{\i}}n Abadi, Ashish Agarwal, Paul Barham, Eugene Brevdo, Zhifeng Chen,
  Craig Citro, Gregory~S. Corrado, Andy Davis, Jeffrey Dean, Matthieu Devin,
  Sanjay Ghemawat, Ian~J. Goodfellow, Andrew Harp, Geoffrey Irving, Michael
  Isard, Yangqing Jia, Rafal J{\'{o}}zefowicz, Lukasz Kaiser, Manjunath Kudlur,
  Josh Levenberg, Dan Man{\'{e}}, Rajat Monga, Sherry Moore, Derek~Gordon
  Murray, Chris Olah, Mike Schuster, Jonathon Shlens, Benoit Steiner, Ilya
  Sutskever, Kunal Talwar, Paul~A. Tucker, Vincent Vanhoucke, Vijay Vasudevan,
  Fernanda~B. Vi{\'{e}}gas, Oriol Vinyals, Pete Warden, Martin Wattenberg,
  Martin Wicke, Yuan Yu, and Xiaoqiang Zheng.
\newblock Tensorflow: Large-scale machine learning on heterogeneous distributed
  systems.
\newblock {\em CoRR}, abs/1603.04467, 2016.

\bibitem{Agethen2019DeepMC}
Sebastian Agethen and Winston~H. Hsu.
\newblock Deep multi-kernel convolutional lstm networks and an attention-based
  mechanism for videos.
\newblock {\em IEEE Transactions on Multimedia}, 22:819--829, 2019.

\bibitem{Arandjelovic2013AllAV}
Relja Arandjelovic and Andrew Zisserman.
\newblock All about vlad.
\newblock {\em 2013 IEEE Conference on Computer Vision and Pattern
  Recognition}, pages 1578--1585, 2013.

\bibitem{Baccouche2010ActionCI}
Moez Baccouche, Franck Mamalet, Christian Wolf, Christophe Garcia, and Atilla
  Baskurt.
\newblock Action classification in soccer videos with long short-term memory
  recurrent neural networks.
\newblock In {\em ICANN}, 2010.

\bibitem{Baughman2019DetectionOT}
Aaron~K. Baughman, Eduardo Morales, Gary Reiss, Nancy~M Greco, Stephen Hammer,
  and Shiqiang Wang.
\newblock Detection of tennis events from acoustic data.
\newblock In {\em MMSports '19}, 2019.

\bibitem{Bridgeman2019MultiPerson3P}
Lewis Bridgeman, Marco Volino, Jean-Yves Guillemaut, and Adrian Hilton.
\newblock Multi-person 3d pose estimation and tracking in sports.
\newblock In {\em CVPR Workshops}, 2019.

\bibitem{Buch2017SSTST}
Shyamal Buch, Victor Escorcia, Chuanqi Shen, Bernard Ghanem, and Juan~Carlos
  Niebles.
\newblock Sst: Single-stream temporal action proposals.
\newblock {\em 2017 IEEE Conference on Computer Vision and Pattern Recognition
  (CVPR)}, pages 6373--6382, 2017.

\bibitem{Cai2018TemporalHA}
Zixi Cai, Helmut Neher, Kanav Vats, David~A. Clausi, and John~S. Zelek.
\newblock Temporal hockey action recognition via pose and optical flows.
\newblock In {\em CVPR Workshops}, 2018.

\bibitem{Cioppa2018ABA}
Anthony Cioppa, Adrien Deli{\`e}ge, and Marc~Van Droogenbroeck.
\newblock A bottom-up approach based on semantics for the interpretation of the
  main camera stream in soccer games.
\newblock {\em 2018 IEEE/CVF Conference on Computer Vision and Pattern
  Recognition Workshops (CVPRW)}, pages 1846--184609, 2018.

\bibitem{Cioppa2019ACL}
Anthony Cioppa, Adrien Deli{\`e}ge, Silvio Giancola, Bernard Ghanem, Marc~Van
  Droogenbroeck, Rikke Gade, and Thomas~B. Moeslund.
\newblock A context-aware loss function for action spotting in soccer videos.
\newblock {\em ArXiv}, abs/1912.01326, 2019.

\bibitem{Cioppa2019ARTHuSAR}
Anthony Cioppa, Antoine Deli{\'e}ge, Maxime Istasse, Christophe~De
  Vleeschouwer, and Marc~Van Droogenbroeck.
\newblock Arthus: Adaptive real-time human segmentation in sports through
  online distillation.
\newblock In {\em CVPR Workshops}, 2019.

\bibitem{Corscadden2018DevelopingAT}
Jack Corscadden, Ross Eastman, Reece Echelberger, Connor Hagan, Clark Kipp,
  Erik Magnusson, G. Muller, Stephen Adams, James Valeiras, and William~T.
  Scherer.
\newblock Developing analytical tools to impact u.va. football performance.
\newblock {\em 2018 Systems and Information Engineering Design Symposium
  (SIEDS)}, pages 249--254, 2018.

\bibitem{Dai2017TemporalCN}
Xiyang Dai, Bharat Singh, Guyue Zhang, Larry~S. Davis, and Yan~Qiu Chen.
\newblock Temporal context network for activity localization in videos.
\newblock {\em 2017 IEEE International Conference on Computer Vision (ICCV)},
  pages 5727--5736, 2017.

\bibitem{Darczy2013FisherKF}
B{\'a}lint~Zolt{\'a}n Dar{\'o}czy, Andr{\'a}s~A. Bencz{\'u}r, and Lajos
  R{\'o}nyai.
\newblock Fisher kernels for image descriptors: a theoretical overview and
  experimental results.
\newblock 2013.

\bibitem{Statista.market.sports}
Deloitte.
\newblock Global sports market - total revenue from 2005 to 2017 (in billion
  u.s. dollars).
\newblock In \textit{Statista - The Statistics Portal}, 2020. Retrieved March
  9, 2020, from
  \href{https://www.statista.com/statistics/370560/worldwide-sports-market-revenue/}{https://www.statista.com/statistics/370560/worldwide-sports-market-revenue/}.

\bibitem{Statista.market.sports.european}
Deloitte.
\newblock Market size of the european football market from 2006/07 to 2015/16
  (in billion euros).
\newblock In \textit{Statista - The Statistics Portal}, 2020. Retrieved March
  9, 2020, from
  \href{https://www.statista.com/statistics/261223/european-soccer-market-total-revenue/}{https://www.statista.com/statistics/261223/european-soccer-market-total-revenue/}.

\bibitem{Statista.market.sports.european.bigfive}
Deloitte.
\newblock Revenue of the biggest (big five*) european soccer leagues from
  1996/97 to 2017/18 (in million euros).
\newblock In \textit{Statista - The Statistics Portal}, 2020. Retrieved March
  9, 2020, from
  \href{https://www.statista.com/statistics/261218/big-five-european-soccer-leagues-revenue/}{https://www.statista.com/statistics/261218/big-five-european-soccer-leagues-revenue/}.

\bibitem{Statista.market.sports.european.top}
Deloitte.
\newblock Revenue of the top european soccer leagues (big five*) from 2006/07
  to 2017/18 (in billion euros).
\newblock In \textit{Statista - The Statistics Portal}, 2020. Retrieved March
  9, 2020, from
  \href{https://www.statista.com/statistics/261225/top-european-soccer-leagues-big-five-revenue/}{https://www.statista.com/statistics/261225/top-european-soccer-leagues-big-five-revenue/}.

\bibitem{imagenet}
Jia Deng, Wei Dong, Richard Socher, Li-Jia Li, Kai Li, and Fei-Fei Li.
\newblock Imagenet: a large-scale hierarchical image database.
\newblock pages 248--255, 06 2009.

\bibitem{DOrazio2010ARO}
Tiziana D'Orazio and Marco Leo.
\newblock A review of vision-based systems for soccer video analysis.
\newblock {\em Pattern Recognit.}, 43:2911--2926, 2010.

\bibitem{Ekin2003AutomaticSV}
Ahmet Ekin, A.~Murat Tekalp, and Rajiv Mehrotra.
\newblock Automatic soccer video analysis and summarization.
\newblock {\em IEEE transactions on image processing : a publication of the
  IEEE Signal Processing Society}, 12 7:796--807, 2003.

\bibitem{Farin2003RobustCC}
Dirk Farin, Susanne Krabbe, Peter H.~N. de With, and Wolfgang Effelsberg.
\newblock Robust camera calibration for sport videos using court models.
\newblock In {\em IS\&T/SPIE Electronic Imaging}, 2003.

\bibitem{Feichtenhofer2016SpatiotemporalRN}
Christoph Feichtenhofer, Axel Pinz, and Richard~P. Wildes.
\newblock Spatiotemporal residual networks for video action recognition.
\newblock {\em ArXiv}, abs/1611.02155, 2016.

\bibitem{Felsen2017WhatWH}
Panna Felsen, Pulkit Agrawal, and Jitendra Malik.
\newblock What will happen next? forecasting player moves in sports videos.
\newblock {\em 2017 IEEE International Conference on Computer Vision (ICCV)},
  pages 3362--3371, 2017.

\bibitem{Gade2015AudioVisualCO}
Rikke Gade, Mohamed Abou-Zleikha, Mads~Gr{\ae}sb{\o}ll Christensen, and
  Thomas~B. Moeslund.
\newblock Audio-visual classification of sports types.
\newblock {\em 2015 IEEE International Conference on Computer Vision Workshop
  (ICCVW)}, pages 768--773, 2015.

\bibitem{Gao2017TURNTT}
Jiyang Gao, Zhenheng Yang, Chen Sun, Kan Chen, and Ramakant Nevatia.
\newblock Turn tap: Temporal unit regression network for temporal action
  proposals.
\newblock {\em 2017 IEEE International Conference on Computer Vision (ICCV)},
  pages 3648--3656, 2017.

\bibitem{audioset}
Jort~F. Gemmeke, Daniel P.~W. Ellis, Dylan Freedman, Aren Jansen, Wade
  Lawrence, R.~Channing Moore, Manoj Plakal, and Marvin Ritter.
\newblock Audio set: An ontology and human-labeled dataset for audio events.
\newblock In {\em Proc. IEEE ICASSP 2017}, New Orleans, LA, 2017.

\bibitem{Giancola_2018_CVPR_Workshops}
Silvio Giancola, Mohieddine Amine, Tarek Dghaily, and Bernard Ghanem.
\newblock Soccernet: A scalable dataset for action spotting in soccer videos.
\newblock In {\em The IEEE Conference on Computer Vision and Pattern
  Recognition (CVPR) Workshops}, June 2018.

\bibitem{Girdhar2017ActionVLADLS}
Rohit Girdhar, Deva Ramanan, Abhinav Gupta, Josef Sivic, and Bryan~C. Russell.
\newblock Actionvlad: Learning spatio-temporal aggregation for action
  classification.
\newblock {\em 2017 IEEE Conference on Computer Vision and Pattern Recognition
  (CVPR)}, pages 3165--3174, 2017.

\bibitem{DBLP:journals/corr/HeZRS15}
Kaiming He, Xiangyu Zhang, Shaoqing Ren, and Jian Sun.
\newblock Deep residual learning for image recognition.
\newblock {\em CoRR}, abs/1512.03385, 2015.

\bibitem{Heilbron2017SCCSC}
Fabian~Caba Heilbron, Wayner Barrios, Victor Escorcia, and Bernard Ghanem.
\newblock Scc: Semantic context cascade for efficient action detection.
\newblock {\em 2017 IEEE Conference on Computer Vision and Pattern Recognition
  (CVPR)}, pages 3175--3184, 2017.

\bibitem{10.1162/neco.1997.9.8.1735}
Sepp Hochreiter and J\"{u}rgen Schmidhuber.
\newblock Long short-term memory.
\newblock {\em Neural Comput.}, 9(8):1735–1780, Nov. 1997.

\bibitem{Homayounfar2017SportsFL}
Namdar Homayounfar, Sanja Fidler, and Raquel Urtasun.
\newblock Sports field localization via deep structured models.
\newblock {\em 2017 IEEE Conference on Computer Vision and Pattern Recognition
  (CVPR)}, pages 4012--4020, 2017.

\bibitem{Huang2006SemanticAO}
Chung-Lin Huang, Huang-Chia Shih, and Chung-Yuan Chao.
\newblock Semantic analysis of soccer video using dynamic bayesian network.
\newblock {\em IEEE Transactions on Multimedia}, 8:749--760, 2006.

\bibitem{Istasse2019AssociativeEF}
Maxime Istasse, Julien Moreau, and Christophe~De Vleeschouwer.
\newblock Associative embedding for team discrimination.
\newblock In {\em CVPR Workshops}, 2019.

\bibitem{Jgou2010AggregatingLD}
Herv{\'e} J{\'e}gou, Matthijs Douze, Cordelia Schmid, and Patrick P{\'e}rez.
\newblock Aggregating local descriptors into a compact image representation.
\newblock {\em 2010 IEEE Computer Society Conference on Computer Vision and
  Pattern Recognition}, pages 3304--3311, 2010.

\bibitem{Jiang2019OptimizingTL}
Wei Jiang, Juan Camilo~Gamboa Higuera, Baptiste Angles, Weiwei Sun, Mehrsan
  Javan, and Kwang~Moo Yi.
\newblock Optimizing through learned errors for accurate sports field
  registration.
\newblock {\em ArXiv}, abs/1909.08034, 2019.

\bibitem{Kamble2017BallTI}
Paresh~R. Kamble, A.~G. Keskar, and K.~M. Bhurchandi.
\newblock Ball tracking in sports: a survey.
\newblock {\em Artificial Intelligence Review}, pages 1--51, 2017.

\bibitem{Khan2019VisualRO}
Abdullah Khan, Loris Bozzato, Luciano Serafini, and Beatrice Lazzerini.
\newblock Visual reasoning on complex events in soccer videos using answer set
  programming.
\newblock In {\em GCAI}, 2019.

\bibitem{Khan2018SoccerED}
Abdullah Khan, Beatrice Lazzerini, Gaetano Calabrese, and Luciano Serafini.
\newblock Soccer event detection.
\newblock In {\em ICIP 2018}, 2018.

\bibitem{Komorowski2019FootAndBallIP}
Jacek Komorowski, Grzegorz Kurzejamski, and Grzegorz Sarwas.
\newblock Footandball: Integrated player and ball detector.
\newblock {\em ArXiv}, abs/1912.05445, 2019.

\bibitem{Liu2019MultiScaleBC}
Haijun Liu, Shiguang Wang, Wen Wang, and Jian Cheng.
\newblock Multi-scale based context-aware net for action detection.
\newblock {\em IEEE Transactions on Multimedia}, 22:337--348, 2019.

\bibitem{Lu2019PantiltzoomSF}
Jikai Lu, Jianhui Chen, and James~J. Little.
\newblock Pan-tilt-zoom slam for sports videos.
\newblock {\em ArXiv}, abs/1907.08816, 2019.

\bibitem{Manafifard2017ASO}
M. Manafifard, Hamid Ebadi, and Hamid~Abrishami Moghaddam.
\newblock A survey on player tracking in soccer videos.
\newblock {\em Comput. Vis. Image Underst.}, 159:19--46, 2017.

\bibitem{Martnez2019ActionRW}
Brais Mart{\'i}nez, Davide Modolo, Yuanjun Xiong, and Joseph Tighe.
\newblock Action recognition with spatial-temporal discriminative filter banks.
\newblock {\em 2019 IEEE/CVF International Conference on Computer Vision
  (ICCV)}, pages 5481--5490, 2019.

\bibitem{Merler2019AutomaticCO}
Michele Merler, Khoi-Nguyen~C. Mac, Dhiraj Joshi, Quoc-Bao Nguyen, Stephen
  Hammer, John Kent, Jinjun Xiong, Minh~N. Do, Joshua~R. Smith, and
  Rog{\'e}rio~Schmidt Feris.
\newblock Automatic curation of sports highlights using multimodal excitement
  features.
\newblock {\em IEEE Transactions on Multimedia}, 21:1147--1160, 2019.

\bibitem{DBLP:journals/corr/MiechLS17}
Antoine Miech, Ivan Laptev, and Josef Sivic.
\newblock Learnable pooling with context gating for video classification.
\newblock {\em CoRR}, abs/1706.06905, 2017.

\bibitem{Nagel2015MetaFV}
Markus Nagel and Jan~C. van Gemert.
\newblock Meta fisher vector for event recognition in generative encoded visual
  streams.
\newblock 2015.

\bibitem{Nguyen2015STAPSA}
Tam~V. Nguyen, Zheng Song, and Shuicheng Yan.
\newblock Stap: Spatial-temporal attention-aware pooling for action
  recognition.
\newblock {\em IEEE Transactions on Circuits and Systems for Video Technology},
  25:77--86, 2015.

\bibitem{Oskouie2012MultimodalFE}
Payam Oskouie, Sara Alipour, and Amir-Masoud Eftekhari-Moghadam.
\newblock Multimodal feature extraction and fusion for semantic mining of
  soccer video: a survey.
\newblock {\em Artificial Intelligence Review}, 42:173--210, 2012.

\bibitem{Pan2019ForegroundFV}
Yongsheng Pan, Yong Xia, and Dinggang Shen.
\newblock Foreground fisher vector: Encoding class-relevant foreground to
  improve image classification.
\newblock {\em IEEE Transactions on Image Processing}, 28:4716--4729, 2019.

\bibitem{Pei2016TemporalAM}
Wenjie Pei, Tadas Baltrusaitis, David M.~J. Tax, and Louis-Philippe Morency.
\newblock Temporal attention-gated model for robust sequence classification.
\newblock {\em 2017 IEEE Conference on Computer Vision and Pattern Recognition
  (CVPR)}, pages 820--829, 2016.

\bibitem{Peng2019TwoStreamCL}
Yuxin Peng, Yunzhen Zhao, and Junchao Zhang.
\newblock Two-stream collaborative learning with spatial-temporal attention for
  video classification.
\newblock {\em IEEE Transactions on Circuits and Systems for Video Technology},
  29:773--786, 2019.

\bibitem{Philbin2008LostIQ}
James Philbin, Ondřej Chum, Michael Isard, Josef Sivic, and Andrew Zisserman.
\newblock Lost in quantization: Improving particular object retrieval in large
  scale image databases.
\newblock {\em 2008 IEEE Conference on Computer Vision and Pattern
  Recognition}, pages 1--8, 2008.

\bibitem{Ren2005FootballVS}
Reede Ren and Joemon~M. Jose.
\newblock Football video segmentation based on video production strategy.
\newblock In {\em ECIR}, 2005.

\bibitem{10.1145/3347318.3355524}
Melissa Sanabria, Sherly, Fr\'{e}d\'{e}ric Precioso, and Thomas Menguy.
\newblock A deep architecture for multimodal summarization of soccer games.
\newblock In {\em Proceedings Proceedings of the 2nd International Workshop on
  Multimedia Content Analysis in Sports}, MMSports ’19, page 16–24, New
  York, NY, USA, 2019. Association for Computing Machinery.

\bibitem{Sarkar2019GenerationOB}
Saikat Sarkar, Amlan Chakrabarti, and Dipti~Prasad Mukherjee.
\newblock Generation of ball possession statistics in soccer using minimum-cost
  flow network.
\newblock In {\em CVPR Workshops}, 2019.

\bibitem{Shou2016TemporalAL}
Zheng Shou, Dongang Wang, and Shih-Fu Chang.
\newblock Temporal action localization in untrimmed videos via multi-stage
  cnns.
\newblock {\em 2016 IEEE Conference on Computer Vision and Pattern Recognition
  (CVPR)}, pages 1049--1058, 2016.

\bibitem{vgg}
Karen Simonyan and Andrew Zisserman.
\newblock Very deep convolutional networks for large-scale image recognition.
\newblock {\em arXiv 1409.1556}, 09 2014.

\bibitem{ec1}
Anastasios Skarlatidis, Georgios Paliouras, Alexander Artikis, and George
  Vouros.
\newblock Probabilistic event calculus for event recognition.
\newblock {\em ACM Transactions on Computational Logic}, 12 2014.

\bibitem{Tang2019DeepFF}
Peng Tang, Xinggang Wang, Baoguang Shi, Xiang Bai, Wenyu Liu, and Zhuowen Tu.
\newblock Deep fishernet for image classification.
\newblock {\em IEEE Transactions on Neural Networks and Learning Systems},
  30:2244--2250, 2019.

\bibitem{Tang2019VideoSC}
Xiaolin Tang.
\newblock Video sequence classification using spatio-temporal features.
\newblock 2019.

\bibitem{Tavassolipour2014EventDA}
Mostafa Tavassolipour, Mahmood Karimian, and Shohreh Kasaei.
\newblock Event detection and summarization in soccer videos using bayesian
  network and copula.
\newblock {\em IEEE Transactions on Circuits and Systems for Video Technology},
  24:291--304, 2014.

\bibitem{Theagarajan2018SoccerWH}
Rajkumar Theagarajan, Federico Pala, Xiu Zhang, and Bir Bhanu.
\newblock Soccer: Who has the ball? generating visual analytics and player
  statistics.
\newblock {\em 2018 IEEE/CVF Conference on Computer Vision and Pattern
  Recognition Workshops (CVPRW)}, pages 1830--18308, 2018.

\bibitem{Thinh2019AVT}
Nguyen~Hong Thinh, Hoang~Minh Son, Chu Thi~Phuong Dzung, Vu~Quang Dzung, and
  Luu~Manh Ha.
\newblock A video-based tracking system for football player analysis using
  efficient convolution operators.
\newblock {\em 2019 International Conference on Advanced Technologies for
  Communications (ATC)}, pages 149--154, 2019.

\bibitem{Thomas2017ComputerVF}
Graham~A. Thomas, Rikke Gade, Thomas~B. Moeslund, Peter Carr, and Adrian
  Hilton.
\newblock Computer vision for sports: Current applications and research topics.
\newblock {\em Comput. Vis. Image Underst.}, 159:3--18, 2017.

\bibitem{Tran2019VideoCW}
Du Tran, Heng Wang, Lorenzo Torresani, and Matt Feiszli.
\newblock Video classification with channel-separated convolutional networks.
\newblock {\em 2019 IEEE/CVF International Conference on Computer Vision
  (ICCV)}, pages 5551--5560, 2019.

\bibitem{Tsunoda2017FootballAR}
Takamasa Tsunoda, Yasuhiro Komori, Masakazu Matsugu, and Tatsuya Harada.
\newblock Football action recognition using hierarchical lstm.
\newblock {\em 2017 IEEE Conference on Computer Vision and Pattern Recognition
  Workshops (CVPRW)}, pages 155--163, 2017.

\bibitem{10.1145/3347318.3355526}
Francesco Turchini, Lorenzo Seidenari, Leonardo Galteri, Andrea Ferracani,
  Giuseppe Becchi, and Alberto Del~Bimbo.
\newblock Flexible automatic football filming and summarization.
\newblock In {\em Proceedings Proceedings of the 2nd International Workshop on
  Multimedia Content Analysis in Sports}, MMSports ’19, page 108–114, New
  York, NY, USA, 2019. Association for Computing Machinery.

\bibitem{Huda2018EstimatingTN}
Noor ul Huda, Kasper~H. Jensen, Rikke Gade, and Thomas~B. Moeslund.
\newblock Estimating the number of soccer players using simulation-based
  occlusion handling.
\newblock {\em 2018 IEEE/CVF Conference on Computer Vision and Pattern
  Recognition Workshops (CVPRW)}, pages 1905--190509, 2018.

\bibitem{Vats2019TwoStreamAR}
Kanav Vats, Helmut Neher, David~A. Clausi, and John~S. Zelek.
\newblock Two-stream action recognition in ice hockey using player pose
  sequences and optical flows.
\newblock {\em 2019 16th Conference on Computer and Robot Vision (CRV)}, pages
  181--188, 2019.

\bibitem{Wang2018FastAA}
Jinzhuo Wang, Wenmin Wang, and Wen Gao.
\newblock Fast and accurate action detection in videos with motion-centric
  attention model.
\newblock {\em IEEE Transactions on Circuits and Systems for Video Technology},
  30:117--130, 2018.

\bibitem{wyscout}
WyScout.
\newblock Wyscout.
\newblock In \textit{WyScout}, 2020. Retrieved March 9, 2020, from
  \href{https://wyscout.com/}{https://wyscout.com/}.

\bibitem{Yang2019ExploringFS}
Ke Yang, Xiaolong Shen, Peng Qiao, Shijie Li, Dong sheng Li, and Yong Dou.
\newblock Exploring frame segmentation networks for temporal action
  localization.
\newblock {\em J. Vis. Commun. Image Represent.}, 61:296--302, 2019.

\bibitem{Yao2016RobustMC}
Qiang Yao, Keisuke Nonaka, Hiroshi Sankoh, and Sei Naito.
\newblock Robust moving camera calibration for synthesizing free viewpoint
  soccer video.
\newblock {\em 2016 IEEE International Conference on Image Processing (ICIP)},
  pages 1185--1189, 2016.

\bibitem{Yaparla2019ANF}
Ganesh Yaparla, Allaparthi Sriteja, Sai~Krishna Munnangi, and Garimella~Rama
  Murthy.
\newblock A novel framework for fine grained action recognition in soccer.
\newblock In {\em IWANN}, 2019.

\bibitem{Ye2019TwoStreamCN}
W. Ye, Jingjng Cheng, Feng Yang, and Yikai Xu.
\newblock Two-stream convolutional network for improving activity recognition
  using convolutional long short-term memory networks.
\newblock {\em IEEE Access}, 7:67772--67780, 2019.

\bibitem{Yu2018FineGrainedVC}
Huanyu Yu, Shuo Cheng, Bingbing Ni, Minsi Wang, Jian Zhang, and Xiaokang Yang.
\newblock Fine-grained video captioning for sports narrative.
\newblock {\em 2018 IEEE/CVF Conference on Computer Vision and Pattern
  Recognition}, pages 6006--6015, 2018.

\bibitem{Yudistira2020CorrelationNS}
Novanto Yudistira and Takio Kurita.
\newblock Correlation net: Spatiotemporal multimodal deep learning for action
  recognition.
\newblock {\em Signal Process. Image Commun.}, 82:115731, 2020.

\bibitem{Zecha2019RefiningJL}
Dan Zecha, Moritz Einfalt, and Rainer Lienhart.
\newblock Refining joint locations for human pose tracking in sports videos.
\newblock In {\em CVPR Workshops}, 2019.

\bibitem{Zhang2019AnAM}
Kailai Zhang, Ji Wu, Xiaofeng Tong, and Yumeng Wang.
\newblock An automatic multi-camera-based event extraction system for real
  soccer videos.
\newblock {\em Pattern Analysis and Applications}, pages 1--13, 2019.

\end{thebibliography}
}

\end{document}